\pdfoutput=1

\documentclass[11pt]{article}

\usepackage[final]{coling}

\usepackage{times}
\usepackage{latexsym}

\usepackage[T1]{fontenc}

\usepackage[utf8]{inputenc}

\usepackage{microtype}

\usepackage{inconsolata}

\usepackage{graphicx}
\usepackage{amsmath}
\usepackage{multirow}
\usepackage{array}
\usepackage{makecell}
\usepackage{booktabs}
\usepackage{hyperref}
\usepackage{enumitem}

\newcommand{\propainsight}{\textit{PropaInsight}}
\newcommand{\propagaze}{\textit{PropaGaze}}
\newcommand{\ptcgaze}{\textit{PTC-Gaze}}
\newcommand{\ruwagaze}{\textit{RUWA-Gaze}}
\newcommand{\politifactgaze}{\textit{Politifact-Gaze}}

\newcount\Comments   
\definecolor{purple}{rgb}{1,0,1}
\newcommand{\kibitz}[2]{\ifnum\Comments=1\textcolor{#1}{#2}\fi}

\title{PropaInsight: Toward Deeper Understanding of Propaganda\\ in Terms of Techniques, Appeals, and Intent}

\author{
    \textbf{Jiateng Liu\textsuperscript{1, \thanks{These authors contribute to this work equally.}}},
    \textbf{Lin Ai\textsuperscript{2, \footnotemark[1]}},
    \textbf{Zizhou Liu\textsuperscript{2}},
    \textbf{Payam Karisani\textsuperscript{1}},
    \textbf{Zheng Hui\textsuperscript{2}},
    \\ 
    \textbf{May Fung\textsuperscript{1}},
    \textbf{Preslav Nakov\textsuperscript{3}},
    \textbf{Julia Hirschberg\textsuperscript{2}},
    \textbf{Heng Ji\textsuperscript{1}}
    \\
    \textsuperscript{1}University of Illinois Urbana-Champaign
    \hspace{1em}
    \textsuperscript{2}Columbia University
    \\
    \textsuperscript{3}Mohamed bin Zayed University of Artificial Intelligence
    \\
    \{jiateng5, hengji\}@illinois.edu,
    \{lin.ai, julia\}@cs.columbia.edu
}

\begin{document}
\maketitle

\begin{abstract}

Propaganda plays a critical role in shaping public opinion and fueling disinformation. While existing research primarily focuses on identifying propaganda techniques, it lacks the ability to capture the broader motives and the impacts of such content. To address these challenges, we introduce \textbf{{\propainsight}}, a conceptual framework grounded in foundational social science research, which systematically dissects propaganda into techniques, arousal appeals, and underlying intent. {\propainsight} offers a more granular understanding of how propaganda operates across different contexts. Additionally, we present \textbf{{\propagaze}}, a novel dataset that combines human-annotated data with high-quality synthetic data generated through a meticulously designed pipeline.

Our experiments show that off-the-shelf LLMs struggle with propaganda analysis, but training with {\propagaze} significantly improves performance.  
Fine-tuned Llama-7B-Chat achieves 203.4\% higher text span IoU in technique identification and 66.2\% higher BertScore in appeal analysis compared to 1-shot GPT-4-Turbo. Moreover, {\propagaze} complements limited human-annotated data in data-sparse and cross-domain scenarios, showing its potential for comprehensive and generalizable propaganda analysis.\footnote{{\propagaze} and code are available at this GitHub repository: \href{https://github.com/Lumos-Jiateng/PropaInsight}{https://github.com/Lumos-Jiateng/PropaInsight}.} 

\end{abstract}
\section{Introduction}
\begin{figure*}[ht]
    \centering
    \includegraphics[width=0.85\textwidth]{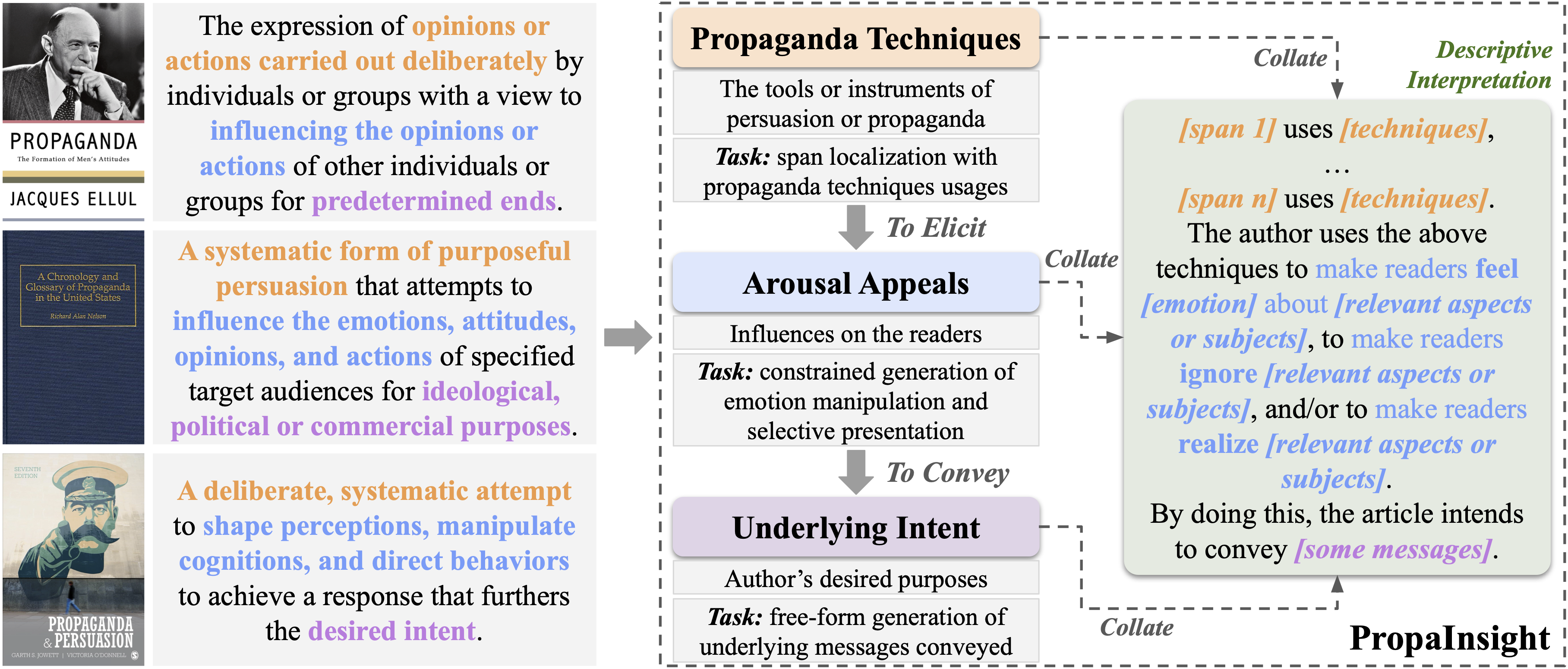}
    \caption{We abstract key elements of propaganda from social science literature. A comprehensive propaganda frame includes the techniques employed, the appeals evoked in readers, and the author's underlying intent.}
    \vspace{-0.2cm}
    \label{fig:propainsight}
\end{figure*}

In an era of unbounded digital information, the deliberate dissemination of propaganda has proliferated, shaping public opinion and influencing political events~\cite{stanley2015propaganda}. Propaganda is also a key component of disinformation, where false information is intentionally crafted and distributed to deceive or mislead~\cite{da2020prta}. Detecting and analyzing propaganda is essential to maintain the integrity of public discourse and to ensure that individuals make informed, unbiased decisions~\cite{da2020prta}.

Most current research on propaganda detection focuses on identifying and categorizing the specific techniques used to persuade the audience~\cite{da2019fine,martino2020semeval}. However, simply recognizing these techniques does not fully capture the motives behind the propaganda or its broader impact. As noted by~\citet{hobbs2020mind,lord2021strengthen}, understanding the sources and the intentions behind the information is critical to effectively combat disinformation. Therefore, there is a growing need for moving beyond simply recognizing techniques to deepen  understanding of propaganda.

Moreover, existing studies often rely on expert annotations because non-expert annotators struggle with separating personal biases from their assessments of propaganda~\cite{da2019fine}. This dependence on expert annotations leads to relatively small datasets, which may be insufficient for training large, generalizable models, and limits their applicability in cross-domain contexts where propaganda usage varies. For instance, strategies in military content can differ greatly from such in political content, highlighting the need for broader, more diverse datasets.

To address these challenges, we build on foundational social science research on propaganda~\cite{nelson1997chronology,jowett2018propaganda,ellul2021propaganda} and identify three key elements behind propaganda attempts: \textit{propaganda techniques}, \textit{arousal appeals}, and \textit{underlying intent}. Consequently, we introduce a new conceptual framework, {\propainsight}, that systematically analyzes these elements. Additionally, we leverage the strong context understanding ability of large language models (LLMs) to generate synthetic data, resulting in {\propagaze}, a novel dataset for propaganda analysis. Our motivations are twofold: \textbf{(1)} to develop a comprehensive framework that goes beyond identifying techniques, and \textbf{(2)} to explore the use of synthetic data to supplement limited human-annotated data. Our contributions are as follows:
\begin{itemize}[leftmargin=*,nosep,topsep=0pt]
    \item[1.] We propose {\propainsight}, a conceptual framework for granular and comprehensive propaganda analysis that identifies propaganda techniques, arousal appeals, and underlying intent in news articles.
    \item[2.] We introduce {\propagaze}, a novel dataset for fine-grained propaganda analysis, consisting of a human-annotated news sub-dataset and two high-quality synthetic sub-datasets: one focused on the Russia-Ukraine conflict and one on the political domain. 
    \item[3.] We demonstrate that {\propagaze} enhances LLMs' ability to analyze propaganda within the {\propainsight} framework, paving the way for more nuanced and generalizable propaganda analysis methods.
\end{itemize}
\section{\propainsight: A Propaganda Analysis Framework}
\label{sec:formulation}


We introduce {\propainsight}, a new conceptual framework for comprehensive propaganda analysis. In contrast to previous methods which ignore the underlying purposes and only focus on techniques, {\propainsight} delves into the more subtle and hidden elements of propaganda. Drawing from foundational social science research on propaganda~\cite{nelson1997chronology,jowett2018propaganda,ellul2021propaganda}, we identify three key elements of each propaganda attempt: \textit{propaganda techniques}, \textit{arousal appeals}, and \textit{underlying intent}. As shown in Figure~\ref{fig:propainsight}, for a given article, we first identify and classify the techniques used. We then infer the arousal appeals these techniques evoke, and we further deduce the underlying intent of the article. To ensure interpretability and consistency, we consolidate these elements into a clear, structured natural language paragraph using a descriptive template, as shown in Figure~\ref{fig:propainsight}. Below, we provide a detailed explanation of each element of our proposed framework. 

\paragraph{Propaganda Techniques}
Propaganda techniques are systematic, deliberate strategies used to craft persuasive content~\cite{jowett2012propaganda}. Domain experts typically define these techniques as pre-defined labels like `loaded language'. 
While the specific techniques may vary across different shared tasks~\cite{torok2015symbiotic}, we follow the set of propaganda techniques defined in~\cite{da2019fine}, where each technique can be evaluated intrinsically. The full list of the 16 propaganda techniques we use is provided in Appendix~\ref{app:template}.


\paragraph{Arousal Appeals} Appeals directly influence a reader's emotions, opinions, and actions after consuming propagandistic content~\cite{nelson1997chronology,jowett2012propaganda}. A common propaganda device is to evoke strong emotions, such as hate or fear, in readers \cite{miller1937detect}. Another approach involves selectively presenting evidence and facts to shape the audience's perception \cite{walton1997propaganda,o2004politics}. To capture these effects, we design three templates (detailed in Appendix~\ref{app:template}) that identify the emotions evoked and the aspects readers are guided toward or distracted from while reading an article.

\paragraph{Underlying Intent} Intent represents the ideological, political, or other underlying goal the author seeks to convey or achieve. To handle diverse real-world scenarios, we frame intent prediction as a free-text generation task, similar to approaches used for open intent generation in dialogue systems~\cite{csimcsek2018intent,wagner2022open}. The advantage of this novel formulation in propaganda intent analysis is its flexibility in capturing complex, nuanced intent that predefined labels cannot easily categorize, allowing greater freedom to generate more detailed and context-specific interpretations of intent.  


\paragraph{Propaganda Analysis Task} 


The design of {\propainsight} introduces a new propaganda analysis task: generating a descriptive natural language paragraph explaining the techniques used, the appeals aroused, and the underlying intent. To avoid overlooking individual elements and to simplify evaluation, we divide the task into three sub-tasks:
\begin{itemize}[leftmargin=*,nosep,topsep=0pt]
    \item[1.] \textbf{\textit{Propaganda Technique Identification:}} Detect the spans where propaganda techniques are applied and which specific technique(s) correspond to each span, following prior task settings~\cite{martino2020semeval,martino2020survey}.
    \item[2.] \textbf{\textit{Appeal Analysis:}} Generate the descriptions of emotions and feelings evoked using a template-based approach (see Appendix~\ref{app:template}).
    \item[3.] \textbf{\textit{Intent Analysis:}} Generate a free-form explanation of the article's underlying intent.
\end{itemize}
\section{\propagaze: A Dataset for Systematically Analyzing Propaganda}
\label{sec:data creation}


\begin{figure*}[tbh]
    \centering
    \includegraphics[width=1\textwidth]{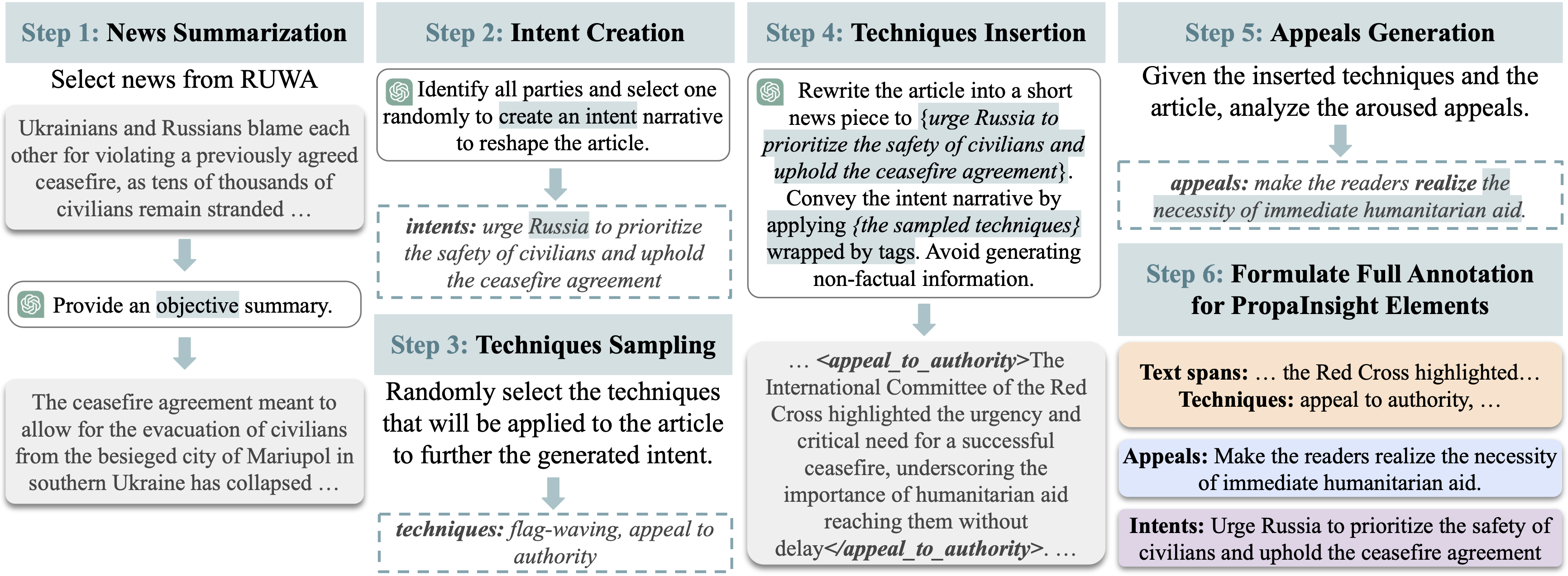}
    \caption{Partially controlled data generation pipeline: We first collect real-world news articles and derive an objective summary to extract events. Then we generate event-based intent, and randomly sample specific propaganda techniques to insert into the event descriptions. Lastly, we generate appeals from a reader's perspective, aiming at making the appeals grounded to the text.}
    \vspace{-0.2cm}
    \label{fig:pipeline}
\end{figure*}

Existing propaganda datasets~\cite{martino2020semeval,heppell-etal-2023-analysing} primarily focus on identifying propaganda techniques and their associated text spans, but lack insights into appeal and intent. We introduce {\propagaze}, a new dataset specifically designed for comprehensive propaganda analysis, consisting of three sub-datasets: {\ptcgaze}, {\ruwagaze}, and {\politifactgaze}.

\subsection{\ptcgaze: Human-Annotated Dataset}
\label{subsec:ptcgaze}
{\ptcgaze} builds on the existing PTC dataset~\cite{martino2020semeval}, which includes human-written news articles annotated for propaganda techniques and spans. We reannotate this dataset by hiring human annotators to label appeals and intent independently. For appeals, annotators review propaganda-containing sentences along with their context and describe the feelings evoked. To reduce cognitive load, we provide GPT-4 generated candidate annotations for assistance. Annotators then evaluate whether the generated candidates accurately reflect their interpretations and reactions, and if not, they rewrite the descriptions based on the template in \S~\ref{sec:formulation} and Appendix~\ref{app:template}. 
For intent, annotators read the full article and infer its underlying intent in a single free-form sentence, and we leave the multi-intent scenarios for future work.
We used Label Studio to design the annotation interface, which is later shown in Appendix~\ref{app:dataset}. Two professional annotators from Kitware Inc. are in charge of the annotation task. We only have one annotator for each annotation task so no agreement rate is computed. As shown inTable~\ref{tab:dataset_stats}, this annotated sub-dataset contains 79 articles, with an average of 12.77 propaganda techniques per article. Additional information, data examples, and analysis of the annotation quality are given in Appendix~\ref{app:dataset}. 

\begin{table}[tbh]
\centering
\resizebox{0.48\textwidth}{!}
{
    \begin{tabular}{lccc}
    \toprule
    \textit{Sub-Dataset $\downarrow$} & \textbf{\#Articles} & \makecell[c]{\textbf{Avg. Article}\\ \textbf{Length (words)}} & \makecell[c]{\textbf{Avg. Propa}\\ \textbf{Usage / Article}} \\
    \cmidrule{2-4}
    \ptcgaze & 79 & 885.16 & 12.77 \\
    \ruwagaze & 497 & 266.04 & 3.46 \\
    \politifactgaze & 593 & 339.05 & 3.47 \\
    \bottomrule
    \end{tabular}
}
\caption{Statistics about the {\propagaze} dataset.}
\vspace{-0.2cm}
\label{tab:dataset_stats}
\end{table}


\subsection{{\ruwagaze} and {\politifactgaze}: Synthetic Datasets}

One limitation of the fully human-annotated dataset is that its usualy expensive, due to the challenging nature of the annotation tasks. 
This makes it insufficient for training large, generalizable models and limits its cross-domain applicability. Sparse data is a common issue in propaganda analysis research. To address this, we leverage LLMs such as LLaMA~\cite{touvron2023llamaopenefficientfoundation} and GPT~\cite{ye2023comprehensive, openai2023gpt4} to synthesize data, using their strong prior knowledge and context understanding. These models have shown effectiveness in data augmentation for tasks like propaganda techniques identification, such as fallacy recognition~\cite{alhindi-etal-2024-large}. These synthetic datasets are created mainly for training and can also serve as silver-standard benchmarks for propaganda analysis.

We construct {\ruwagaze} and {\politifactgaze} using a partially controlled data generation pipeline, as illustrated in Figure \ref{fig:pipeline}. Specifically, {\ruwagaze} is built upon a dataset of real-world news articles focused on the Russia-Ukraine War~\cite{ruwa2023}, while {\politifactgaze} is constructed using the PolitiFact partition of the FakeNewsNet dataset~\cite{shu2020fakenewsnet}.

\paragraph{Data Generation Pipeline} Figure~\ref{fig:pipeline} shows the data creation pipeline. Initially, we use GPT-3.5 to summarize human-written, published news articles and to identify key events and objective facts. These summaries are intended to be objective, as the original articles may reflect various biases that could influence the creation of new propaganda pieces. Following this, we use GPT-3.5 to extract all focal entities 
involved in the events. We then randomly select one entity's 
perspective and set an intent to guide the revision of the article. We also randomly choose a set of propaganda techniques to be inserted into the article, reshaping its narrative. 

Subsequently, we use GPT-4 as an intermediary author to craft intentional propaganda articles based on real-world events by injecting sampled propaganda techniques into an objective summary. We also ask the model to self-analyze the appeals the rewritten article might evoke to ensure alignment with the established intent. Human readers then verify the data quality for any obvious errors. The prompts for each step are provided in Appendix~\ref{app:data_generation_prompt}. 


As illustrated in Table~\ref{tab:dataset_stats}, {\ruwagaze} consists of 497 articles, and {\politifactgaze} consists of 593 articles. 
While we generated moderate data 
 due to the computational cost. We believe the data generation pipeline is generalizable. The language models can be replaced with cheaper or open-source LLMs to reduce costs and, in turn, generate larger-scale datasets. In addition, we identify that these two subsets come from different domains (Military \& War and Politics), and they differ significantly in both content and the use of propaganda techniques.

\section{Experiments}
\label{sec:experiments}

LLMs have strong prior knowledge and advanced context understanding, which makes them ideal for synthesizing propaganda-rich datasets and potentially effective for analyzing propaganda. In this section, we explore three research questions: \textbf{(1)} how off-the-shelf LLMs perform on propaganda analysis, \textbf{(2)} how much the {\propagaze} dataset improves performance when used for training or fine-tuning, and \textbf{(3)} whether propaganda analysis is transferable across domains.

\subsection{Experimental Setup}
\label{sec:4.1}

\paragraph{Sub-Tasks and Metrics} As outlined in \S~\ref{sec:formulation}, {\propainsight} makes it possible to break the propaganda analysis task into three sub-tasks to ensure detailed evaluation and capture key elements:
\begin{itemize}[leftmargin=*,nosep,topsep=0pt]
    \item[1.] \textbf{\textit{Propaganda Techniques Identification:}} 
    We use Intersection over Union (\textbf{IoU}) to measure the overlap between the identified text spans and the actual propaganda spans, and \textbf{F1} scores to evaluate propaganda technique classification, following prior task settings~\cite{martino2020semeval,martino2020survey}.
    \item[2.] \textbf{\textit{Appeals Analysis:}} 
    We evaluate the quality of the generated responses using \textbf{BertScore}~\cite{zhang2019BertScore} to measure semantic similarity.
    \item[3.] \textbf{\textit{Intent Analysis:}} 
    Similarly, we use \textbf{BertScore} for this sub-task.
\end{itemize}

\paragraph{Models} We experiment with the following:
\begin{itemize}[leftmargin=*,nosep,topsep=0pt]
    \item[1.] \textbf{GPT-4-Turbo}: One of the top-performing OpenAI models for various tasks. We use it in both zero-shot and few-shot prompting settings across all sub-tasks. The specific prompts used for each sub-task are given in Appendix~\ref{app:template}.
    \item[2.] \textbf{Llama-7B-Chat}: A popular open-source LLM. Due to its smaller size and relatively worse performance compared to GPT-4-Turbo, we fine-tune it for our sub-tasks. Specifically, we instruction-tune it to predict whether each sentence contains propaganda, and if so, identify the techniques and the appeals used, and predict the article's intent. See Appendix~\ref{app:template} for the fine-tuning prompts. 
    \item[3.] \textbf{Multi-Granularity Neural Networks (MGNN model)}~\cite{da2019fine}: A benchmark method for the propaganda techniques identification sub-task. We train MGNN from scratch for this specific task, as it is not designed for text generation and cannot be applied to the other two sub-tasks. 
\end{itemize}

\paragraph{Data-Rich and Data-Sparse Training Settings}
In real-world scenarios, obtaining a large volume of well-annotated data for analyzing propaganda is challenging, as discussed in \S~\ref{sec:data creation}. 
6tt

For all {\propagaze} sub-datasets, we split the articles into training and testing sets using a 70:30 ratio. {\ptcgaze}, with only 79 articles, represents a data-sparse condition. In contrast, the synthetic sub-datasets, {\ruwagaze} and {\politifactgaze}, contain a total of over 1,000 articles. To simulate data-sparse scenarios with these two sub-datasets, we sample subsets matching the size of the full {\ptcgaze} training set. For data-rich conditions, we use the full training sets of {\ruwagaze} and {\politifactgaze}, reserving one-seventh as the validation set.

\subsection{How Do Off-the-Shelf LLMs Perform on Propaganda Analysis Tasks?}
\label{subsec:baselines}
\begin{table*}[ht]
\centering
\resizebox{0.95\textwidth}{!}
{
    \begin{tabular}{clcccccccc}
    \toprule
    & \textbf{Dataset $\rightarrow$} & \multicolumn{2}{c}{\textbf{\ruwagaze}} & & \multicolumn{2}{c}{\textbf{\politifactgaze}} & & \multicolumn{2}{c}{\textbf{\ptcgaze}} \\
    \textbf{Data Setting $\downarrow$} & \textbf{Model $\downarrow$} & \makecell{Span\\Avg. IoU} & \makecell{Techniques\\Macro F1} & & \makecell{Span\\Avg. IoU} & \makecell{Techniques\\Macro F1} & & \makecell{Span\\Avg. IoU} & \makecell{Techniques\\Macro F1} \\
    \cmidrule{3-4} \cmidrule{6-7} \cmidrule{9-10}
    \multirow{3}{*}{\textbf{\textit{No Training Data}}} & GPT-4-Turbo$_{0s}$ & 0.073 & 0.097 & & 0.152 & 0.226   & & 0.124 & 0.068 \\
    & GPT-4-Turbo$_{1s}$ & 0.132 & 0.145 & & 0.183&  0.269  & & \underline{0.165} & 0.171  \\
    \midrule
    \multirow{2}{*}{\textbf{\textit{Data-Sparse Training}}} & MGNN & 0.089 & 0.139 & & 0.160 & 0.159 & & 0.140 & \textbf{0.206} \\
    & Llama-7B-Chat$_{ft}$ & 0.230 & 0.210 & & 0.253 & 0.281 & & \textbf{0.179} & \underline{0.191} \\
    \midrule
    \multirow{2}{*}{\textbf{\textit{Data-Rich Training}}} & MGNN & \textbf{0.545} & \underline{0.591} & & \textbf{0.449} & \textbf{0.461} & & - & - \\
    & Llama-7B-Chat$_{ft}$ & \underline{0.506} & \textbf{0.607} & & \underline{0.409} & \underline{0.453} & & - & - \\

    \bottomrule
    \end{tabular}
}
\caption{Model performance on the propaganda technique identification sub-task under different training data settings. We report the performance of trained MGNN model and both k-shot ($ks$) and fine-tuned ($ft$) LLMs.}
\label{tab:techniques_results_in_domain}
\end{table*}
\begin{table*}[ht]
\centering
\resizebox{0.95\textwidth}{!}
{
    \begin{tabular}{lcccccccc}
    \toprule
    \textbf{Dataset $\rightarrow$} & \multicolumn{2}{c}{\textbf{\ruwagaze}} & & \multicolumn{2}{c}{\textbf{\politifactgaze}} & & \multicolumn{2}{c}{\textbf{\ptcgaze}} \\
    \textbf{Model $\downarrow$} & \makecell{Appeals\\BertScore} & \makecell{Intents\\BertScore} & & \makecell{Appeals\\BertScore} & \makecell{Intents\\BertScore} & & \makecell{Appeals\\BertScore} & \makecell{Intents\\BertScore} \\
    \cmidrule{2-3} \cmidrule{5-6} \cmidrule{8-9}
    GPT-4-Turbo$_{0s}$ & 0.282 & 0.849 &  & 0.298 &  0.863 & & 0.228 & \underline{0.869} \\
    GPT-4-Turbo$_{1s}$ & \underline{0.324} & \textbf{0.879} &  & \underline{0.345} & \textbf{0.875} & & \textbf{0.331} & \textbf{0.881} \\
    Llama-7B-Chat$_{ft}$ \textit{(Data-Sparse)} & 0.313 &  0.851  & & 0.342 & 0.860 & & \underline{0.249} & 0.843 \\
    Llama-7B-Chat$_{ft}$ \textit{(Data-Rich)} & \textbf{0.612} & \underline{0.861} & & \textbf{0.495} & \underline{0.864} & & - & - \\

    \bottomrule
    \end{tabular}
}
\caption{Model performance on appeal and intent analysis sub-tasks under different training data settings. We report the performance of zero-shot ($0s$) and fine-tuned ($ft$) LLMs.}
\vspace{-0.2cm}
\label{tab:appelas_results_in_domain}
\end{table*}

As shown in Tables~\ref{tab:techniques_results_in_domain} and \ref{tab:appelas_results_in_domain}, \textbf{zero-shot LLMs struggle with propaganda analysis}. For example, in identifying propaganda techniques, zero-shot GPT-4-Turbo underperforms compared to the trained MGNN, even in data-sparse conditions, despite MGNN being much smaller in size. Zero-shot LLMs often struggle to pinpoint sentences containing propaganda. Similarly, in appeal analysis, zero-shot GPT-4-Turbo achieves relatively low BertScores. However, these models perform better at inferring intent, as shown by their stronger performance in the intent analysis sub-task (Table~\ref{tab:appelas_results_in_domain}).

\textbf{Few-shot prompting improves LLM performance in analyzing propaganda elements}. Specifically, in identifying propaganda techniques, one-shot GPT-4-Turbo shows an 80.8\% improvement in average IoU on {\ruwagaze}, a 20.4\% increase on {\politifactgaze}, and a 33.1\% higher IoU on {\ptcgaze} compared to a zero-shot setting. Similarly, in appeal analysis, one-shot GPT-4-Turbo achieves 14.9\% higher BertScore on {\ruwagaze}, 15.8\% higher on {\politifactgaze}, and 45.2\% higher on {\ptcgaze}. In intent analysis, zero-shot GPT-4-Turbo already performs well. The improvements compared to one-shot prompting are minor, with the highest increase being 3.5\% on {\ruwagaze}.

\subsection{How Much Does {\propagaze} Enhance Model Performance?}
\label{subsec:main_results}

\textbf{{\propagaze} substantially improves the overall propaganda analysis performance}, especially in identifying propaganda techniques, under both data-sparse and data-rich training conditions. In the data-sparse setting, fine-tuned LLaMA-7B-Chat outperforms one-shot GPT-4-Turbo, achieving an average of 65.8\% higher text span IoU and 33.7\% higher technique identification F1 score, as shown in Table~\ref{tab:techniques_results_in_domain}. In the data-rich setting, the performance increases even further, with LLaMA-7B-Chat showing 90.9\% higher text span IoU and 125.1\% higher F1 score compared to the data-sparse results.
Table~\ref{tab:appelas_results_in_domain} shows similar improvements in appeals and intent analysis. For the appeals sub-task, data-rich fine-tuning leads to an average 70.1\% increase in BertScore, while for intent analysis there is a smaller 8.5\% gain compared to data-sparse training. This is likely due to the already high baseline performance. These results demonstrate that the synthetic sub-datasets effectively complement the limited human-annotated data, significantly improving the model's performance in analyzing propaganda elements.

We also compare the performance of LLaMA-7B-Chat with the baseline benchmark MGNN on propaganda identification. In the data-sparse setting, fine-tuned LLaMA-7B-Chat substantially outperforms trained MGNN, achieving 158.43\% higher IoU on {\ruwagaze} and 58.1\% higher IoU on PolitiFact-Gaze. However, in data-rich scenarios, MGNN, benefiting from the larger amount of training data, surpassing LLaMA-7B-Chat. This may be due to the fact that smaller models, such as MGNN, can overfit when exposed to excessive training data, while larger LLMs, such as LLaMA-7B-Chat, generalize better in data-sparse conditions. These findings suggest that \textbf{LLMs are more suited for the task with limited training data, while smaller, dedicated models like MGNN could benefit more from the synthetic sub-datasets provided by {\propagaze} in data-rich environments}. This is consistent with the findings of \citet{alhindi-etal-2024-large}. Thus, with sufficient training data, we can implement a pipeline that first localizes and identifies propaganda techniques using MGNN, followed by appeals and intent analysis based on MGNN's output. This approach could potentially enhance the overall quality of the model's output for the entire propaganda analysis task.

\begin{table*}[ht]
\centering
\resizebox{0.95\textwidth}{!}
{
    \begin{tabular}{clcccccccc}
    \toprule
    & \textbf{Eval Dataset $\rightarrow$} & \multicolumn{2}{c}{\textbf{\ruwagaze}} & & \multicolumn{2}{c}{\textbf{\politifactgaze}} & & \multicolumn{2}{c}{\textbf{\ptcgaze}} \\
    \makecell[l]{\textbf{Additional}\\ \textbf{Train Data} $\downarrow$} & \textbf{Model $\downarrow$} & \makecell{Span\\Avg. IoU} & \makecell{Techniques\\Macro F1} & & \makecell{Span\\Avg. IoU} & \makecell{Techniques\\Macro F1} & & \makecell{Span\\Avg. IoU} & \makecell{Techniques\\Macro F1} \\
    \cmidrule{3-4} \cmidrule{6-7} \cmidrule{9-10}
    \multirow{2}{*}{\textbf{\ruwagaze}} & MGNN$_{ft}$ & 0.089 $|$ \textbf{0.545} & 0.139 $|$ 0.591 & & 0.243 $|$ \textbf{0.471} & 0.251 $|$ \textbf{0.475} & & 0.157 $|$ 0.224 & \underline{0.212} $|$ 0.272 \\
    & Llama-7B-Chat$_{ft}$ & 0.230 $|$ \underline{0.506} & 0.210 $|$ \textbf{0.607 }& & \underline{0.262} $|$ 0.379 & \underline{0.274} $|$ 0.418 & & \textbf{0.215} $|$ \underline{0.243} & \textbf{0.220} $|$ 0.258  \\
    \midrule
    \multirow{2}{*}{\textbf{\politifactgaze}} & MGNN$_{ft}$ & \underline{0.246} $|$ 0.456 & \textbf{\textbf{0.281}} $|$ \underline{0.593} & & 0.160 $|$ \underline{0.449} & 0.159 $|$ \underline{0.461} & & \underline{0.203} $|$ \textbf{0.252} & 0.210 $|$ \textbf{0.298} \\
    & Llama-7B-Chat$_{ft}$ & \textbf{0.271} $|$ 0.443 & \underline{0.265} $|$ 0.582 & & 0.253 $|$ 0.409 & \textbf{0.281} $|$ 0.453 & & 0.196 $|$ 0.237 & 0.204 $|$ \underline{0.273} \\
    \midrule
    \multirow{2}{*}{\textbf{\ptcgaze}} & MGNN$_{ft}$ & 0.189 $|$ - - - - & 0.226 $|$ - - - - & & 0.224 $|$ - - - - & 0.237 $|$ - - - - & & 0.140 $|$ - - - - & 0.206 $|$ - - - - \\
    & Llama-7B-Chat$_{ft}$ & 0.215 $|$ - - - - & 0.239 $|$ - - - - & & \textbf{0.265} $|$ - - - - & 0.261 $|$ - - - - & & 0.179 $|$ - - - - & 0.191 $|$ - - - - \\

    \bottomrule
    \end{tabular}
}
\caption{Model performance (\textbf{data-sparse $|$ data-rich}) on the propaganda techniques identification sub-task under cross-domain training. The \textbf{best result} and \underline{runner-up result} are highlighted per column for the data-sparse and data-rich settings, respectively. Diagonal cells show in-domain training only, without cross-domain training, and are included for reference.}
\label{tab:cross_domain_techniques_results}
\end{table*}
\begin{table*}[ht]
\centering
\resizebox{0.95\textwidth}{!}
{
    \begin{tabular}{ccccccccc}
    \toprule
    \textbf{Eval Dataset $\rightarrow$} & \multicolumn{2}{c}{\textbf{\ruwagaze}} & & \multicolumn{2}{c}{\textbf{\politifactgaze}} & & \multicolumn{2}{c}{\textbf{\ptcgaze}} \\
    \textbf{Additional Train Data} $\downarrow$ & \makecell{Appeals\\BertScore} & \makecell{Intents\\BertScore} & & \makecell{Appeals\\BertScore} & \makecell{Intents\\BertScore} & & \makecell{Appeals\\BertScore} & \makecell{Intents\\BertScore} \\
    \cmidrule{2-3} \cmidrule{5-6} \cmidrule{8-9}
    \textbf{\ruwagaze} & 0.313 $|$ \textbf{0.612} & \underline{0.851} $|$ \textbf{0.861} & & \textbf{0.362} $|$ \underline{0.452} & 0.858 $|$ \textbf{0.865} & & \textbf{0.293} $|$ \textbf{0.352} & 0.839 $|$ \underline{0.841} \\
    \textbf{\politifactgaze} & \textbf{0.373} $|$ \underline{0.584} & \textbf{0.855} $|$ \underline{0.860} & & 0.342 $|$ \textbf{0.495} & \underline{0.860} $|$ \underline{0.864} & & \underline{0.267}$|$ \underline{0.310} & \textbf{0.845} $|$ \textbf{0.847} \\
    \textbf{\ptcgaze} & \underline{0.366} $|$ - - - - & 0.848 $|$ - - - - & & \underline{0.350} $|$ - - - - & \textbf{0.863} $|$ - - - - & & 0.249 $|$ - - - - & \underline{0.843} $|$ - - - - \\
    \bottomrule
    \end{tabular}
}
\caption{Fine-tuned Llama-7B-Chat model performance (\textbf{data-sparse $|$ data-rich}) on the appeals and intent analysis sub-tasks under cross-domain training. The \textbf{best result} and \underline{runner-up result} are highlighted per column for the data-sparse and data-rich settings, respectively. Diagonal cells show in-domain training only, without cross-domain training, and are included for reference.}
\vspace{-0.2cm}
\label{tab:cross_domain_appeals_intents_results}
\end{table*}
\subsection{Is Propaganda Analysis Transferable Across Domains?}
\label{subsec:cross_domain}
In the real world, propaganda spans various domains, including military and war, politics, economics, science, environmental issues, and more. Although the specific use of propaganda may differ across these domains, we are particularly interested in determining whether the general patterns of propaganda are transferable between domains. Additionally, high-quality human-annotated data is scarce, prompting us to investigate whether leveraging data from other domains can improve propaganda analysis in a target domain.

As outlined in \S~\ref{sec:data creation}, our dataset consists of three subsets: {\ruwagaze} (military and war), {\politifactgaze} (politics), and {\ptcgaze} (general news). To explore cross-domain transferability, we perform additional training on each target sub-dataset using data from the other two sub-datasets after the in-domain training. For instance, in a data-sparse scenario, if {\ruwagaze} is the target, cross-domain training on {\politifactgaze} involves first training the model on the sparse {\ruwagaze} data, followed by further training with sparse {\politifactgaze} data. In a data-rich scenario, the model is trained on the full in-domain {\ruwagaze} data, then further trained on the entire {\politifactgaze} dataset. The results are presented in Tables~\ref{tab:cross_domain_techniques_results} and \ref{tab:cross_domain_appeals_intents_results}.

In data-sparse settings, we observe that models benefit substantially from incorporating cross-domain data. As shown in Table~\ref{tab:cross_domain_techniques_results}, when evaluated on {\ruwagaze}, models trained on additional data from {\politifactgaze} and {\ptcgaze} achieve higher performance than those trained solely on sparse in-domain data. Specifically, LLaMA-7B-Chat fine-tuned with additional {\politifactgaze} data achieves the highest text span IoU of 0.271, while MGNN trained with additional {\politifactgaze} data reaches the highest technique F1 score of 0.281. This pattern is consistent across other sub-datasets and holds true for appeal analysis as well, as shown in Table~\ref{tab:cross_domain_appeals_intents_results}. This is expected, as models trained in data-sparse conditions tend to benefit from cross-domain data due to the need for a larger pool of training examples. 
Access to additional data from related domains enables models to learn generalized patterns of propaganda usage more effectively, leading to improved performance even on tasks outside of their original training domain.

However, in data-rich scenarios, the benefit of cross-domain training diminishes. For example, as shown in Table~\ref{tab:cross_domain_techniques_results}, models trained on additional {\politifactgaze} data underperform those trained solely on in-domain data when evaluated on {\ruwagaze}. Similarly, when evaluated on {\politifactgaze}, adding {\ruwagaze} data sometimes leads to performance improvements, but the gains are relatively small. This holds for appeal analysis as well, as we can see in Table~\ref{tab:cross_domain_appeals_intents_results}. These results suggest that \textbf{when there is sufficient training data, the quality of the data has a greater impact on performance than its quantity}. We further observe that training on both {\ruwagaze} and {\politifactgaze} improves the performance on the human-annotated {\ptcgaze} across all sub-tasks. While this is partly due to the data-sparse nature of {\ptcgaze}, making extra training samples valuable, it also highlights that our synthetic data effectively complements the limited human-annotated data.

\section{Discussion}

\subsection{Discrepancy between Human-Annotated and Synthetic Datasets}
\label{subsec:discrepancy_human_synthetic_data}
We acknowledge the discrepancy between the synthetic sub-datasets and the human-annotated sub-dataset in {\propagaze}. As shown in Table~\ref{tab:dataset_stats}, the average number of propaganda techniques per article in {\ptcgaze} is 12.77, which is about 3.7 times higher than in the synthetic {\ruwagaze} and {\politifactgaze}. This occurs due to the way we generate the synthetic data, where we inject three propaganda techniques per article, with GPT-4-Turbo sometimes reusing techniques. However, we believe this is less of an issue, as {\ptcgaze} articles are on average 3.3 times longer than those in the other sub-datasets. Moreover, since we treat the injected techniques as silver labels, we have not yet checked whether other sentences in the articles also use propaganda techniques. See the Limitations section for more details. Finally, we note the inherent difference in writing styles between synthetic and human-written articles, which is a common challenge with synthetic datasets.

\subsection{Further Challenges of Propaganda Analysis}
\label{subsec:further_challenges}
We identified that accurately pinpointing the occurrence of propaganda is a major challenge in propaganda analysis. As highlighted in the case study (Appendix~\ref{app: case study}), LLMs often misclassify non-propagandistic sentences as propagandistic, leading to a high false positives rate. This issue may be partially attributed to hallucination or failing to account for subtle contextual differences. Although less frequent, similar errors occur with MGNN, indicating that the problem lies not only in the models themselves, but also in the training methodologies and the underlying algorithms. This underscores the need for improvements in both model development and in the training approaches to better distinguish propagandistic content from neutral text.
\section{Related Work}

\paragraph{Propaganda Detection}
Propaganda detection has long been a focus in both Natural Language Processing, with most work focusing on identifying propaganda usage and specific techniques. Various learning-based approaches have improved performance~\cite{da2019fine,yoosuf2019fine,li2019detection,vorakitphan2022protect} and interpretability~\cite{yu2021interpretable} in detecting propaganda in news articles~\cite{vlad2019sentence,da2019fine,gupta2019neural,yu2021interpretable} and tweets~\cite{vijayaraghavan2022tweetspin,khanday2021detecting,guarino2020characterizing}. Recent efforts have also applied LLMs to this task~\cite{sprenkamp2023large,jones2024detecting}. While these studies focus on identifying propaganda techniques, further research is needed to understand the appeals and intent behind them.

Following the escalation of the Russo-Ukrainian conflict in 2022, research has focused on analyzing propaganda campaigns, particularly from Russia. \citet{chen2023tweets,fung2022weibodataset2022russoukrainian} collected user content and opinions from social media platforms such as X and Weibo, while \citet{golovchenko2022fighting} examined censorship of Ukrainian content on Russian platforms.  
\citet{geissler2023russian} studied pro-Russian sentiment on social media and the role of bots, and \citet{patrona2022snapshots} explored intertextuality and rhetoric in political performances during the war. However, few studies developed frameworks to analyze the specific intent behind propagandistic efforts. \citet{ai2024tweetintent} examined two specific propaganda narrative intentions, but failed short of proposing a generalizable framework for propaganda analysis.

\paragraph{Propaganda Generation}
Compared to propaganda detection, research on propaganda generation is sparse. \citet{zellers2019defending} explores generating propaganda to spread targeted disinformation, while \citet{huang-etal-2023-faking} focuses on incorporating emotional and non-emotional propaganda techniques into generated articles. \citet{goldstein2024persuasive} find that GPT-3 can generate highly persuasive propaganda. Our data generation pipeline goes further by allowing a broader range of propaganda techniques to be inserted into generated articles to evoke specific intent, while allowing for more granular analysis of the appeals behind their use.

\paragraph{User Intent Detection}
Previous methods on intent detection concentrated primarily on understanding user queries in human-machine dialogue systems~\cite{brenes2009survey,liu2019review,weld2022survey}. This research facilitated the construction of more robust search engines and virtual assistants. The similarity of this task to ours is that both tasks require strong natural language understanding. However, detecting user query intent is relatively superficial compared to the intent behind a propaganda tactic, which could be highly concealed and hard to recognize~\cite{jowett2012propaganda}. 


\section{Conclusion and Future Work}

We proposed a comprehensive approach to propaganda analysis that goes beyond simply identifying techniques and addresses the common challenge of obtaining high-quality human-annotated data. We further introduced {\propainsight}, a conceptual framework for granular propaganda analysis that identifies propaganda techniques, arousal appeals, and underlying intent, grounded in foundational social science research. Moreover, we presented {\propagaze}, a novel dataset for fine-grained propaganda analysis that includes both human-annotated and high-quality synthetic sub-datasets. Our experiments showed that models fine-tuned on {\propagaze} outperform one-shot GPT-4-Turbo by a margin. {\propagaze} proved highly beneficial in data-sparse and cross-domain scenarios, serving as an effective complement to limited human-annotated data.

Furthermore, {\propainsight} has broader implications beyond propaganda analysis. It enhances tasks such as disinformation detection~\cite{guess2020misinformation,Ai_Chen_Gong_Guo_Hooshmand_Yang_Hirschberg_2021,huang-etal-2023-faking}, sentiment analysis~\cite{ahmad2019review}, narrative framing~\cite{colley2019strategic,andersen2020islamic}, media bias analysis~\cite{nakov2021fake, zollmann2019bringing}, and social media monitoring~\cite{chaudhari2021propaganda}, offering deeper insights into manipulative content and coordinated disinformation campaigns, making the framework applicable to a wide range of areas.In the future, we plan to expand {\propagaze} into more diverse domains and genres, which will further broaden the scope of propaganda analysis. We will also explore how {\propainsight} can improve downstream applications and contribute to a deeper understanding of propaganda.

\section*{Limitations}

We reflect on the limitations of our work below:

\begin{enumerate}[leftmargin=*,nosep]
    \item Although our dataset, {\propagaze}, is novel and reliable, its size is relatively small due to the computational costs associated with GPT-4 and the high expense of human annotation. Consequently, we are uncertain about the dataset's ability to generalize across a broad range of domains when models are fine-tuned exclusively on it.

    \item While we aimed to include diverse domains and construct a cross-domain dataset, the vast range of real-world scenarios exceeds what we could capture. The extent of the domain gap where propaganda thrives remains unclear, and therefore, the cross-domain performance we tested across our paper might not generalize well under varied conditions.

    \item Despite our careful calibration of the proposed propaganda framework, the real-world responses such as reader engagement and ultimate impact can vary significantly. Personalized appeals may emerge, influencing the effectiveness of propaganda; however, our study did not account for these individual differences. We did not take this into consideration and leave this part for future work.

    \item Although we use a partially controlled pipeline to generate synthetic data and have basic human reviewers skim the content to ensure its overall quality, a more fine-grained review is necessary. Specifically, we need to assess whether the sampled propaganda techniques are contextually appropriate within each article. Additionally, while we treat the injected techniques as silver labels for our experiments, we do not check whether other sentences in the article, beyond those explicitly marked, also employ propaganda techniques. This means that our synthetic sub-datasets have high precision in labeled techniques but have not been evaluated for recall. It is likely that in reshaping the articles, additional sentences may also use propaganda techniques not explicitly labeled. Further evaluation, potentially involving more comprehensive human annotation, is needed for a more granular assessment of the dataset’s quality.

    \item Our research is currently limited to English, which may restrict the generalizability of our findings to other languages. Future work will focus on extending the approach to cross-lingual settings to address this limitation.

    \item We acknowledge that while our framework attempts to model intrinsic aspects of propaganda, the experimental setup simplifies the characterization of propaganda intentions. Specifically, the process of generating and validating intentions relies on annotator feedback rather than leveraging domain-specific intent modeling, which may limit the system’s ability to fully capture nuanced propaganda strategies.

\end{enumerate}

Based on this, we propose several promising future directions to further push for the success of combating misinformation.

\begin{enumerate}[leftmargin=*,nosep]
    \item A larger dataset developed using our propaganda framework could be constructed to further evaluate how synthetic data enhances the misinformation detection capabilities of large language models.

    \item Collect data from various domains where significant domain gaps typically exist, and investigate whether cross-domain data substantially influence the accuracy of models' understanding and detection capabilities.

    \item Consider personalized responses \cite{responseprediction2023,sun2023decodingsilentmajorityinducing,personadb2024} to propaganda. Conduct an in-depth analysis of how propaganda articles truly affect their readers and explore how these effects differ from the author's original perspective. In terms of solutions, a mixture of experts can probably lead to a better result. It is also worth considering solutions with LLM-based agents~\cite{guo2024large}, which typically include multi-round of planning~\cite{liu-etal-2023-language} and interactions~\cite{wang2024mintevaluatingllmsmultiturn} or code execution~\cite{leti2023,yang2024llmwizardcodewand,codeact2024} to work out a reasoning based solution. 

    \item Enhancing and shaping the knowledge of LLMs may help to cultivate a better understanding of propaganda articles from different perspectives. It is promising to combine perspective-based datasets to state-of-the-art knowledge control approaches~\cite {han2024wordembeddingssteerslanguage,liu2024evediteventbasedknowledgeediting,deng2024unkeunstructuredknowledgeediting} to alter the perspective and standpoints of LLMs.

    \item It would also be valuable to explore incorporating this deeper understanding of propaganda techniques, appeals, and intent to enhance situation understanding and improve the comprehensiveness of situation report generation \cite{reddy2024smartbookaiassistedsituationreport}.

    \item We plan to incorporate intent mining approaches that include domain-specific characterizations of propaganda intentions. By integrating techniques such as hierarchical intent modeling, pretrained language models fine-tuned for intent detection, and emotion-informed analysis, we aim to strengthen our framework’s ability to detect and classify propaganda intentions with greater specificity and robustness.

\end{enumerate}

\section*{Ethical Considerations}
Our paper introduces a pipeline capable of generating news articles with strong intent and the potential for propaganda use. While our primary goal is to leverage this synthetic data to combat misinformation, it is important to acknowledge that the same technology could be misused to produce high-quality deceptive news content that could mislead public opinion. Consequently, the use of this pipeline must be approached with caution and safeguarded to prevent exploitation by malicious actors.

\section*{Acknowledgement}
This research was done with funding from the Defense Advanced Research Projects Agency (DARPA) under Contracts No. HR001120C0123 and HR0011-24-3-0325. The views, opinions, and/or findings expressed are those of the authors and should not be interpreted as representing the official views or policies of the Department of Defense or the U.S. Government. We also thank Kitware.Inc. and Rapidata.Inc. for their help in the data annotation process.
\bibliography{custom}

\appendix
\label{appendix}

\clearpage
\section{Details of a Propaganda Frame}
\label{app:propaganda frame}

We list the closed set of propaganda techniques that are used in the paper in Table~\ref{tab:definition}. We also included the full template that we used to describe appeals and intent. Note that (1) The set of propaganda techniques included here can be freely extended with any other techniques. (2) We made the templates for Appeals and intent with a valid rationale, as detailed in Section\S ~\ref{sec:formulation}. However, we are not claiming that this is the optimal template design among all other possible designs. We believe that prompt engineering and further human assessment are necessary for discovering the optimal template for this task. And we leave this part for future work.

\begin{table}[!htbp]
    \centering
    \small
    \begin{tabular}{lm{1.3in}}
    \hline\hline
    \multicolumn{2}{c}{\textbf{Propaganda Technique Set}} \\
    
    \textit{Loaded Language} & \textit{Name Calling / labeling} \\
    \textit{Repetition} & \textit{Obfuscation} \\
    \textit{Doubt} & \textit{Straw man} \\
    \textit{Flag-waving} & \textit{Causal oversimplification} \\
    \textit{Slogans} & \textit{Black-and-white fallacy} \\
    \textit{Appeal to authority} & \textit{thought-terminating cliche} \\
    \textit{Whataboutism} & \textit{Reductio and Hitlerum} \\
    \textit{Smears} & \textit{Glittering Generalities} \\
    \hline\hline
    \multicolumn{2}{c}{\textbf{Appeals Raised by Propaganda Usage}}          \\ 
    \multicolumn{2}{c}{\textit{1. Make the readers feel [Emotion] about [Sth that is related]}} \\
    \multicolumn{2}{c}{\textit{2. Make the readers realize that [Sth that is related]}} \\
    \multicolumn{2}{c}{\textit{3. Make the readers ignore that [Sth that is related]}} \\
    \hline \hline
    \multicolumn{2}{c}{\textbf{Ultimate Intent Generation}}  \\ 
    \multicolumn{2}{c}{\textit{[Anything that is applicable for explaining the intent]}} \\
    \hline\hline
    \end{tabular}
    \caption{The complete formulation for each component during propaganda analysis. The parts marked by `[]' indicates the allowance for free generation.}
    \label{tab:definition}
\end{table}

\section{Data Generation Prompt Templates}
\label{app:data_generation_prompt}
\paragraph{Step1: News Summarization}
\begin{quote}
    \textit{System Message: You are a helpful assistant.}
    
    News: \{news\}

    Provide an objective summary of the news article, ensuring to present information in neutral terms. Avoid using language or phrases that show bias towards either party involved.
\end{quote}

\paragraph{Step 2: Intent Creation}
\begin{quote}
    \textit{System Message: You are a helpful assistant.}
    
    Article: \{article\}

    Identify all parties mentioned in the article. Select one party randomly and create an intent narrative to potentially reshape the article.
\end{quote}

\paragraph{Step 4: Techniques Insertion}
\begin{quote}
    \textit{System Message: You are a skilled journalist, proficient in composing short brief news pieces.}
    
    Article: \{article\}

    Rewrite the article into a short news piece to \{intent\}. Convey the intent narrative effectively by applying the following rhetorical tactics, once or more as needed. The revision must be concise, with a clear emphasis on using these tactics to communicate the intended message. Avoid generating non-factual information.

    1. \{appeal tactic 1\}\\
    Example: \{appeal tactic 1 example\}

    2. \{appeal tactic 2\}\\
    Example: \{appeal tactic 2 example\}

    3. \{appeal tactic 3\}\\
    Example: \{appeal tactic 3 example\}
\end{quote}

\paragraph{Step 4: Appeals Generation}
\begin{quote}
    \textit{System Message: You are a helpful assistant that identifies how the writer of a news article wants the readers of the article to feel after reading some sentences.}

    In this task, the input will be a news article, then some sentence in the article will be provided and you need to identify how the specific sentence raises appeals among the readers, the propaganda tactics used in these sentences will also be proivded as a hint. Also remeber that your response should be aware of the main goal of the whole article. For each sentence, you only need to output a sentence describing the feelings in one of the following two templates:

    Make the readers feel [Positive \& Negative Emotions] about [Something that is related]\\
    Make the readers realize/Ignore [Something that is related]

    Here is an example indicating the input and output format:
    
    Input:
    News article: \{article\}

    Sentence: \{first sentence\}\\
    Tactic: \{the tactic that is used in the sentence\}

    Sentence: \{second sentence\}\\
    Tactic: \{the tactic that is used in the sentence\}

    ...

    Output:

    [1] Your response for the first sentence: Make the readers feel [Positive \& Negative Emotions] about [Something that is related]
    
    [2] Your response for the second sentence: Make the readers realize/Ignore [Something that is related]
    
    ...
    
    Now let's begin!
    
    Now given the following news article:
\end{quote}

\section{Templates and Prompts We Used for Propaganda Analysis}\label{app:template}

This section describes each generation and prompt used in this paper. While these prompts could be enhanced through prompt engineering and additional human evaluation, we use them here as proof of concept.

\paragraph{Template for Composing the Predicted Elements into a Descriptive Sentence}
We use the following template to compose the predicted elements into a descriptive sentence as the final output for the propaganda analysis task: 
\begin{quote}
    This article uses propaganda to further its author's ultimate intent of \textit{\{The ultimate intent that is predicted by the model\}}. Specifically, the author uses \textit{\{The first identified propaganda technique\}} in the sentence: "\textit{\{The first sentence that is identified to use propaganda\}}" to make the readers \textit{\{The first appeal that is raised among the readers\}}. The author also uses \textit{\{The second identified propaganda technique\}} in the sentence: "\textit{\{The second sentence that is identified to use propaganda\}}" to make the readers \textit{\{The second appeal that is raised among the readers\}}...
\end{quote}

\paragraph{Prompt Template for the Language Models to Analyze Propaganda in a Zero-shot Manner} We use the following prompt to encourage language models to correctly predict the elements with in a propaganda frame, this prompt also enables simple parsing to obtain the results.

\begin{quote}
News article: \{The news article that needs to be analyzed\}

Given the news article above, you should detect the major intent of the article. The intent is conveyed by using certain tactics and raise appeals in some text spans. You are also going to output all the text spans and the corresponding tactics and appeals. 

The tactics that maybe used are listed here: loaded language, flag waving, slogans, repetition, straw man, red herring, whataboutism, obfuscation, causal oversimplification, false dilemma, thought terminating cliche, appeal to authority, bandwagon, glittering generalities, name calling, doubt, smears, reducito ad hitlerum

You should also formulate the generated appeals in the following format, choose one of the following template to fill in the appeals: 

Make the readers feel [Some Emotion] about [Something that is related]
Make the readers realize about [Something that is related]
Make the readers ignore [Something that is related]

Your should firstly output the ultimate intent, then sequentially output all the text spans within the original article that contains tacic and appeals related to this intent and the corresponding tactics and appeals. You should only output one appeal for each text span. Here is an example:

\{Ultimate intent\} The intent detected \{Ultimate intent\}

\{Text Span\} Text Span 1 \{Text Span\}
\{Tactic\} Tactic 1 \{Tactic\}
\{Appeal\} Appeal 1 \{Appeal\}

\{Text Span\} Text Span 2 \{Text Span\}
\{Tactic\} Tactic 2 \{Tactic\}
\{Appeal\} Appeal 1 \{Appeal\}

...

Now, output your answer with the given News article! 
    
\end{quote}

\paragraph{Template for Instruction Tuning with Llama2-Chat-7B on Tactics}

\begin{quote}

User: Is the sentence below using propaganda techniques? Answer with [Yes] [Propaganda Technique] or [No] [None], candidate techniques are: {Providing the candidate techniques} Sentence: \{The sentence that needs to be identified. \}

Assistant: \{The Templated Answers\}

\end{quote}

\paragraph{Template for Instruction Tuning with Llama2-Chat-7B on Appeals}

\begin{quote}

What is the appeal that the author tries to arouse in the following sentence ? Answer with 'Make the readers ...', Sentence: \{The sentence that needs to be identified. \}

Assistant: \{The Templated Answers\}

\end{quote}

\paragraph{Template for Instruction Tuning with Llama2-Chat-7B on intent}

What is the intent that the author tries to convey in the following article ? Answer with a paragraph of intent. Article: \{The article that needs detection. \}

Assistant: \{The Answers\}

\section{Details of the {\propagaze} Dataset}
\label{app:dataset}
As introduced in Section \S ~\ref{sec:data creation}, the {\propagaze} dataset comprises three subsets: {\ruwagaze}, {\politifactgaze}, and {\ptcgaze}. More details and data examples are provided in this section.

\paragraph{{\ruwagaze}} The {\ruwagaze} dataset is constructed focusing on the Ukraine-Russia War. The original news dataset was from~\cite{ruwa2023}. After human verification on the construction steps, we keep 497 articles, with each article having an average of 3.46 times propaganda usage. We provide an example piece of data from the constructed {\ruwagaze} below, the article has three times of propaganda usage: 

\textbf{Generated News Article:} In light of recent intelligence reports from Ukraine detailing a covert Russian operation to recover classified remains from a sunken cruiser in Crimea, the international call for transparency has never been louder. The Moskva missile cruiser, a symbol of strength and resilience, met its fate in the Black Sea on April 14, igniting a flurry of claims and counterclaims between Ukraine and Russia regarding the circumstances of its sinking. Ukraine has steadfastly maintained that the cruiser was struck by two of its Neptune missiles, a claim seemingly bolstered by the Pentagon's confirmation, while Russia vehemently denies such an event. In this critical moment of geopolitical tension, the need for concrete evidence from Ukraine to substantiate its claims cannot be overstated. As we stand united in our pursuit of truth and justice, it becomes imperative for us to rally around the call for transparency, ensuring that every claim made is backed by irrefutable proof. This is not just about a sunken ship; it's about maintaining the delicate balance of peace and preventing any further escalation that could lead our brothers and sisters into an unwarranted conflict. Some may argue, why this focus on Ukraine's need to present evidence when there are other pressing issues at hand that demand our attention. However, this moment offers a unique opportunity to address underlying issues of trust and accountability in a world already beset by misinformation and conflict. By embracing a posture of openness, Ukraine can lead by example, joining the chorus of nations that have chosen the path of transparency and responsibility. Such a move would not only vindicate Ukraine's claims but also strengthen international confidence in its commitment to integrity and truth. As the situation develops, the world watches closely. The presentation of solid evidence will be a pivotal step in resolving the current standoff, soothing tensions, and charting a course towards resolution and understanding between nations.

\textbf{The sentence that uses propaganda 1:} As we stand united in our pursuit of truth and justice, it becomes imperative for us to rally around the call for transparency, ensuring that every claim made is backed by irrefutable proof.

\textbf{Technique annotation 1:} Flag-waving

\textbf{Appeal annotation 1:} Make the readers feel positive about the demonstration of unity and transparency.

\textbf{The sentence that uses propaganda 2:} Some may argue, why this focus on Ukraine's need to present evidence when there are other pressing issues at hand that demand our attention.

\textbf{Technique annotation 2:} Whataboutism

\textbf{Appeal annotation 2:} Make the readers ignore the distraction of other issues and focus on Ukraine's need to present evidence.

\textbf{The sentence that uses propaganda 3:} By embracing a posture of openness, Ukraine can lead by example, joining the chorus of nations that have chosen the path of transparency and responsibility.

\textbf{Technique annotation 3:} Bandwagon

\textbf{Appeal annotation 3:} Make the readers realize the opportunity Ukraine has to set a strong example of transparency and responsibility.
 
\textbf{Intent annotation:} This article urges Ukraine to provide concrete evidence to support their claim that the Moskva missile cruiser was hit by Ukrainian-made Neptune missiles, in order to maintain transparency and prevent further escalation of tensions with Russia.

\paragraph{{\politifactgaze}} Similar to the {\ruwagaze} dataset, this subset is also constructed with our partially controlled generation pipeline. This dataset is based on the PolitiFact partition of the FakeNewsNet dataset~\cite{shu2020fakenewsnet} with a focus on political status across countries. We keep 593 generated articles with with each article having an average of 3.47 times propaganda usage. 
We provide an example piece of data from the constructed {\politifactgaze} below, the article has four times of propaganda usage:

\paragraph{Generated News Article:} Alabama Attorney General John Simmons has taken a bold stand for truth and justice by filing charges against Mary Lynne Davies, the accuser of Roy Moore, who now faces allegations of falsification, a glaring first-degree misdemeanor. In a shocking twist, Davies, a Democrat, has been exposed as a fabricator of claims against Moore, specifically relating to an alleged incident that supposedly occurred when she was 14 and he was in his early 30s. Legal experts confirm the charges could result in a year of imprisonment and \$10,000 in fines for Davies. This sordid tale of deceit came to a head after the yearbook inscription that Moore purportedly wrote in the 1970s was definitively debunked as a forgery. Davies, in a desperate bid for attention, concocted a story so heinous, it has now backfired, forcing her into hiding following her release on \$500 bail. Despite the unraveling of her narrative, her attorney maintains, without evidence, that she is a victim, not a perpetrator of falsehood. The root of this entire scandal lies solely in the vindictive actions of an individual who sought to manipulate public opinion for political gain\u2014disregarding the serious harm inflicted on Roy Moore's reputation and life. Moore's attorney has extended an olive branch, stating that Moore harbors no ill will and remains committed to championing women's rights, a testament to his character and integrity. In a move that restores faith in the justice system, Attorney General Simmons is not only holding Davies accountable but is also investigating other dubious claims against Moore. This encompasses looking into malicious allegations made by employees of a mall, who labeled him with reprehensible titles without a shred of proof. Through the exposing of these falsehoods and the pursuit of accountability, Roy Moore stands vindicated. Legal and moral authorities alike have rallied to his defense, recognizing the travesty that nearly cost a steadfast advocate for American values and family principles his reputation and career. The pursuit of justice for Moore sends a strong message against the weaponization of false accusations in political warfare.

\textbf{The sentence that uses propaganda 1:} Legal experts confirm the charges could result in a year of imprisonment and \$10,000 in fines for Davies.

\textbf{Technique annotation 1:} Appeal to authority

\textbf{Appeal annotation 1:} Make the readers feel anxious and sympathetic towards Davies about her potential legal consequences.

\textbf{The sentence that uses propaganda 2:} This sordid tale of deceit came to a head after the yearbook inscription that Moore purportedly wrote in the 1970s was definitively debunked as a forgery.

\textbf{Technique annotation 2:} Loaded Language

\textbf{Appeal annotation 2:} Make the readers feel disgusted about the forgery that Moore was accused of.

\textbf{The sentence that uses propaganda 3:} The root of this entire scandal lies solely in the vindictive actions of an individual who sought to manipulate public opinion for political gain\u2014disregarding the serious harm inflicted on Roy Moore's reputation and life.

\textbf{Technique annotation 3:} Causal Oversimplification

\textbf{Appeal annotation 3:} Make the readers feel anger and resentment about the sole individual who manipulated public opinion.

\textbf{The sentence that uses propaganda 4:} Legal and moral authorities alike have rallied to his defense, recognizing the travesty that nearly cost a steadfast advocate for American values and family principles his reputation and career

\textbf{Technique annotation 4:} Appeal to authority

\textbf{Appeal annotation 4:} Make the readers feel relief and satisfaction about the support Moore is receiving from legal and moral authorities.

\textbf{Intent annotation:} This article urges Ukraine to provide concrete evidence to support their claim that the Moskva missile cruiser was hit by Ukrainian-made Neptune missiles, in order to maintain transparency and prevent further escalation of tensions with Russia.

\paragraph{{\ptcgaze}:} The {\ptcgaze} subset is constructed based on the propaganda techniques corpus~\cite{martino2020semeval}, the propaganda articles within the PTC dataset are from real-world news articles and the usage of propaganda together with propaganda techniques annotation is done by human annotators. To simulate the propaganda frame usage in the real world, we further hire human annotators from kitware to annotate each propaganda usage with further information of appeals, and conclude the article with intent. To alleviate the annotation burden, we firstly let GPT-4 models to generate a synthetic annotation, and then let human annotators to just the generated parts of this synthetic annotation and further rewrite into their own annotations. We collected 79 long articles with each article has an average number of 12.77 times of propaganda usage. We notice that in the real-world propaganda corpus, the times of propaganda usage for each article can be far exceeding that of synthetic data. We attribute this difference to a domain gap existing between synthetic articles and real articles. We give an example of the annotated article from {\ptcgaze} as below: 

\paragraph{Real-World News Article:} Ex-Sailor Pardoned By Trump Says He's SUING Obama And Comey  A former Navy sailor, who is one of five people to receive a pardon from President Donald Trump, is planning to file a lawsuit against Obama administration officials.  Kristian Saucier, who served a year in federal prison for taking photos of classified sections of the submarine on which he worked, says he was subject to unequal protection by the law.  Saucier said that he realizes he had erred in taking the photos, which he said he wanted to show only to his family to show them where he worked.  He has also lashed out at Obama officials, saying that his prosecution was politically motivated, prompted by sensitivity about classified information amid the scandal involving Clinton's emails.  According to Fox News, Saucier argues that the same officials who sought out punishment to Saucier for his actions chose to be lenient with Hillary Clinton in her use of a private email server and mishandling of classified information.  Saucier's lawyer, Ronald Daigle, told Fox News on Monday that the lawsuit, which he expects to file soon in Manhattan, will name the U.S. Department of Justice, former FBI Director James Comey and former President Barack Obama as defendants, among others.  Saucier, who lives in Vermont, pleaded guilty in 2016 to taking photos inside the USS Alexandria while it was stationed in Groton, Connecticut, in 2009.  He said he only wanted service mementos, but federal prosecutors argued he was a disgruntled sailor who had put national security at risk by taking photos showing the submarine's propulsion system and reactor compartment and then obstructed justice by destroying a laptop and camera. Fox News They interpreted the law in my case to say it was criminal, Saucier told Fox News, referring to prosecuting authorities in his case, but they didn't prosecute Hillary Clinton.  Hillary is still walking free.  Two guys on my ship did the same thing and weren't treated as criminals.  We want them to correct the wrong. Daigle said that a notice about the pending lawsuit was sent to the Department of Justice and others included in it in December.  There is usually a six-month period that must elapse before the lawsuit actually is actually filed.  My case was usually something handled by military courts, he said.  They used me as an example because of [the backlash over] Hillary Clinton, he continued, alleging his life was ruined for political reasons.  With a pardon, there's no magic wand that gets waved and makes everything right, Saucier said, But I try to stay positive and look forward. Saucier has had cars repossessed and is in debt due to the loss of income after having a felony on his record.  The government actively destroyed his life and made it all but impossible for his family to get back on track.  But Hillary Clinton is running around free, to this day.  And that is what Saucier is so burnt about, with good reason.  

\textbf{The sentence that uses propaganda 1:} Fox News: They interpreted the law in my case to say it was criminal, Saucier told Fox News, referring to prosecuting authorities in his case, but they didn't prosecute Hillary Clinton.

\paragraph{Technique annotation 1:} Whataboutism

\paragraph{Appeal annotation 1:} Make the readers feel indignant about the contrasting legal treatments toward Saucier and Hillary Clinton.

\textbf{The sentence that uses propaganda 2:} Two guys on my ship did the same thing and weren't treated as criminals.

\paragraph{Technique annotation 2:} Whataboutism

\paragraph{Appeal annotation 2:} Make the readers feel unjust about the inequality in punitive measures for similar actions.

\textbf{The sentence that uses propaganda 3:} They used me as an example because of [the backlash over] Hillary Clinton, he continued, alleging his life was ruined for political reasons.

\paragraph{Technique annotation 3:} Causal oversimplification

\paragraph{Appeal annotation 3:} Make the readers feel sympathetic towards Saucier's inopportune life circumstances allegedly resulting from political motivations.

\paragraph{Intent Annotation:} The news intends to inform the public about Kristian Saucier's plans to sue Obama administration officials.

\paragraph{Analysis for Annotation Quality}

\begin{figure*}[ht]
    \centering
    \includegraphics[width=1.0\textwidth]{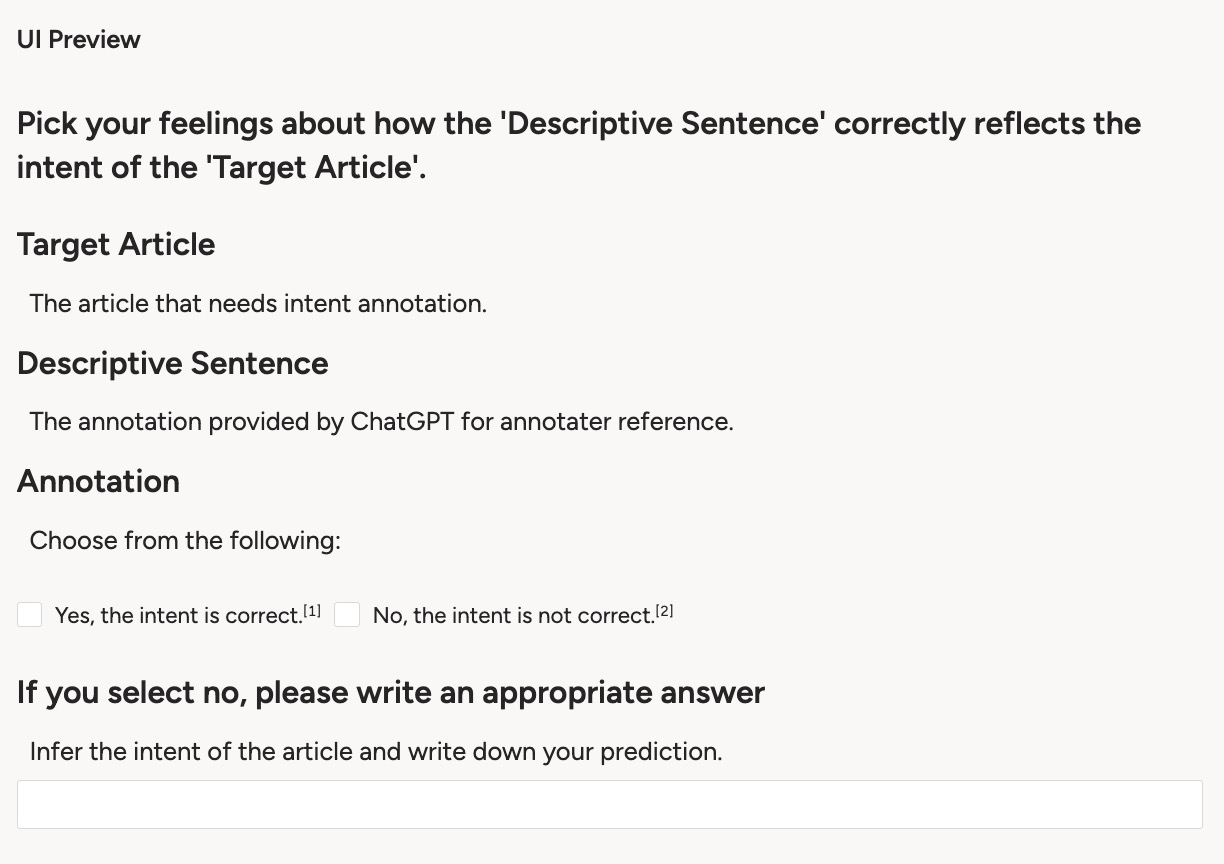}
    \caption{The user interface we used in Label Studio to annotate intent based on a given article.}
    \label{fig:uiintent}
\end{figure*}

\begin{figure*}[ht]
    \centering
    \includegraphics[width=1.0\textwidth]{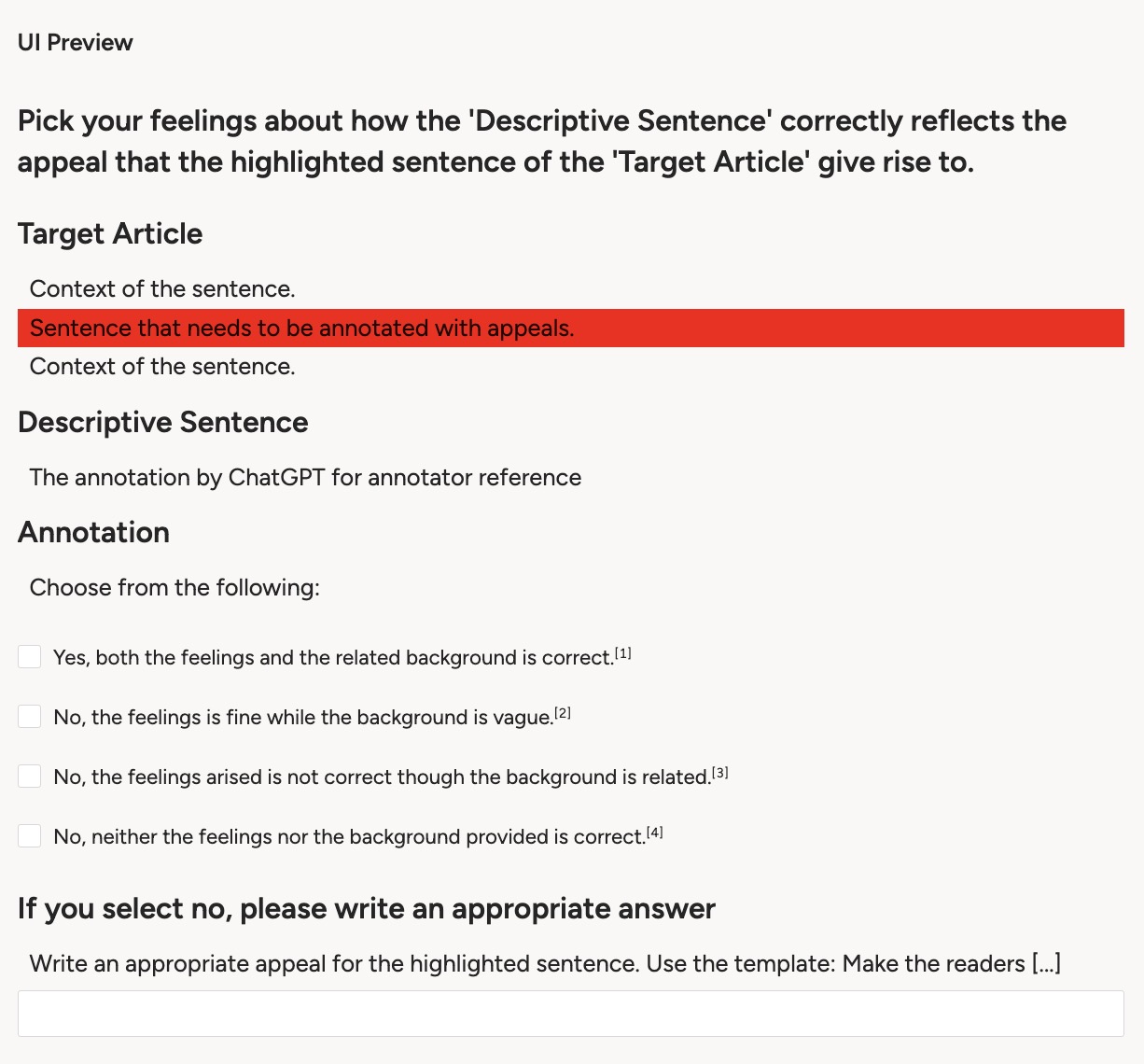}
    \caption{The user interface we used in Label Studio to annotate appeals based on a context. The highlighted part will be the sentence to be annotated, while other parts of 'Target Article' provide related context.}
    \label{fig:uiappeal}
\end{figure*}

We provide an analysis of the annotation quality of our {\ptcgaze} dataset. We used Label Studio for design the annotation interface. We present the user interface design of the intent annotation and appeal annotation tasks in Figure~\ref{fig:uiintent} and Figure~\ref{fig:uiappeal}. Two professional annotators from Kitware.Inc is in charge of the annotation task. Annotators choose to utilize the candidate annotation generated by GPT-4 under 59.8\% annotated intent data points and 75.1\% annotated appeal data points. This demonstrates the high quality of GPT4-provided annotation in terms of appeals and intent, further enhanced our points in \S~\ref{subsec:further_challenges}.
\section{Experimental Details}
\label{app:experimental details}

We provide experimental details for fine-tuning with Llama-Chat-7B and MGNN. For Llama-Chat-7B model, we used the LMFlow~\cite{diao2023lmflow} framework for fine-tuning. We used four A100 GPUs for training, we set the learning rate to 0.00002 and batch size to 4. We tune the model for 3 epochs with our training data. During inference, we always set the inference temperature of the Llama-Chat-7B model to 1.0. For GPT-4-turbo, we used the default temperature for generation. In terms of tuning MGNN, we set batch size to 16 as MGNN takes a smaller memory space, and we set the learning rate to 0.00003. We then tune the model for 20 epochs.

\section{Case Study: Bottleneck of Propaganda Analysis}
\label{app: case study}

As discussed in \S~\ref{subsec:further_challenges}, we find that the bottleneck of propaganda analysis lies in identifying the correct propagandistic sentences. In this section, we give a case study on LLMs doing propaganda analysis to explain the cause further.

\paragraph{Input Example Data}
In a riveting instance of journalism that pierced through the veil of political spin, Fox News host Shepard Smith launched into a fervent condemnation of Donald Trump Jr.'s misleading explanations about his meeting with a Russian lawyer. During a segment that left audiences grappling with notions of truth and integrity, Smith vociferously questioned the incessant \"lies\" and \"deception\" that seem to shroud the Trump administration's dealings, implying the audience's own complacency in the face of such deceit unless they demand accountability. Smith's critique, grounded in an urgent plea for transparency, resonated strongly in an era where allegations of Russian collusion loom over the presidential election\u2014a matter of paramount significance under investigation by multiple congressional committees and a special counsel. \"Why all these lies? Why is it lie after lie after lie? If you're clean, come on clean,\" Smith implored on \"Shepard Smith Reporting,\" emphasizing the sheer implausibility of the evolving narrative woven by Trump Jr. and, by extension, the administration. In a moment that laid bare the discomforting truths surrounding this saga, Smith's acrimonious outburst was underscored by an on-air exchange with fellow anchor Chris Wallace, whose own speechlessness served as a testament to the gravity of Smith's statements. The interaction, a compelling dramatization of the inner turmoil gripping the nation, amplified the weight of Smith's words as he navigated the treacherous waters of political discourse. Yet, amid this pursuit of clarity and honesty, voices emerged calling for Smith's removal from the network\u2014a diversion that starkly contrasts the core issue at hand: the integrity of democratic institutions and the transparency of those in power. Smith's unapologetic defense of the press earlier in the year, where he rebuked claims against CNN as \"not fake news,\" further cements his role as a stalwart advocate for journalistic integrity in the face of political adversity. As the narrative of Donald Trump Jr.'s Russian rendezvous unfolds, Shepard Smith's impassioned critique serves as a poignant reminder of the media's critical role in dissecting the complex web of political narratives, urging the public to remain vigilant, question narratives, and demand nothing short of the truth.

\paragraph{Ground Truth Answers}

We list the ground truth propaganda identification and their related appeals and intent below: 

\begin{quote}
    \textbf{Sentence:} During a segment that left audiences grappling with notions of truth and integrity, Smith vociferously questioned the incessant \"lies\" and \"deception\" that seem to shroud the Trump administration's dealings, implying the audience's own complacency in the face of such deceit unless they demand accountability.

    \textbf{Technique:} loaded language

    \textbf{Appeal:} Make the readers realize the serious implications of Smith's statements, as emphasized by Chris Wallace's speechlessness.

    \textbf{Sentence:} In a moment that laid bare the discomforting truths surrounding this saga, Smith's acrimonious outburst was underscored by an  on-air exchange with fellow anchor Chris Wallace, whose own speechlessness served as a testament to the gravity of Smith's statements.

    \textbf{Technique:} appeal to authority

    \textbf{Appeal:} Make the readers realize the serious implications of Smith's statements, as emphasized by Chris Wallace's speechlessness.

    \textbf{Sentence:} Yet, amid this pursuit of clarity and honesty, voices emerged calling for Smith's removal from the network\u2014a diversion that starkly contrasts the core issue at hand: the integrity of democratic institutions and the transparency of those in power.

    \textbf{Technique:} red herring

    \textbf{Appeal:} Make the readers ignore calls for Smith's removal and focus instead on the integrity of democratic institutions and transparency of those in power.

    \textbf{Intent:} Highlight Fox News host Shepard Smith's criticism of Donald Trump Jr.'s \"lies\" and \"deception\" regarding his meeting with a Russian lawyer, questioning the ongoing narrative of dishonesty and secrecy within the Trump administration.
\end{quote}

\textbf{Predictions from LLMs}

\begin{quote}
    \textbf{Sentence:} In a riveting instance of journalism that pierced through the veil of political spin, Fox News host Shepard Smith launched into a fervent condemnation of Donald Trump Jr.'s misleading explanations about his meeting with a Russian lawyer.

    \textbf{Technique:} loaded language

    \textbf{Appeal:} Make the readers feel inspired by Shepard Smith's courage to confront political deceit.

    \textbf{Sentence:} Smith vociferously questioned the incessant \"lies\" and \"deception\" that seem to shroud the Trump administration's dealings.

    \textbf{Technique:} repetition

    \textbf{Appeal:}  Make the readers realize about the persistent dishonesty in the Trump administration.

    \textbf{Sentence:} Implying the audience's own complacency in the face of such deceit unless they demand accountability.

    \textbf{Technique:} false dilemma

    \textbf{Appeal:} Make the readers feel guilty for any lack of action against political deception.

    \textbf{Sentence:} ``Why all these lies? Why is it lie after lie after lie? If you're clean, come on clean,''

    \textbf{Technique:} repetition

    \textbf{Appeal:} Make the readers feel outraged at the continuous lies by encouraging them to demand honesty.

    \textbf{Sentence:} Smith's critique, grounded in an urgent plea for transparency,

    \textbf{Technique:} appeal to authority

    \textbf{Appeal:} Make the readers trust in the necessity of transparency for integrity.

    \textbf{Sentence:} the sheer implausibility of the evolving narrative woven by Trump Jr. and, by extension, the administration.

    \textbf{Technique:} doubt

    \textbf{Appeal:} Make the readers skeptical about any explanations from the Trump administration.

    \textbf{Sentence:} voices emerged calling for Smith's removal from the network\u2014a diversion that starkly contrasts the core issue at hand: the integrity of democratic institutions and the transparency of those in power.

    \textbf{Technique:} red herring

    \textbf{Appeal:} Make the readers ignore the calls for Smith's removal and focus on the larger issues of integrity and transparency. 

    \textbf{Sentence:} " Smith's unapologetic defense of the press earlier in the year, where he rebuked claims against CNN as \"not fake news,\" "

    \textbf{Technique:} appeal to authority

    \textbf{Appeal:} Make the readers feel supportive of Smith's stance on defending press integrity. 

    \textbf{Sentence:} urging the public to remain vigilant, question narratives, and demand nothing short of the truth.

    \textbf{Technique:} glittering generalities

    \textbf{Appeal:} Make the readers realize about their role in ensuring political honesty and transparency. 

    \textbf{Intent:} The intent detected is to criticize the Trump administration's lack of transparency and integrity, particularly focusing on Donald Trump Jr., while highlighting Shepard Smith's role as a defender of truth and journalistic integrity.
    
\end{quote}

\paragraph{Analysis for the Presented Result}

In the example above, LLMs are making reasonable predictions in terms of 
general intent, with Bertscore over 0.9, however, the model gives poor predictions on the sentences which are propagandistic. Though only three sentences are identified as propagandistic, the LLM predicts there is nine. Despite the techniques included covers the ground truth techniques, its low grounding rate causes these predictions to be useless. 

\end{document}